# Prediction of Hemolysis Tendency of Peptides using a Reliable Evaluation Method

Ali Raza
Dept. of Computer and Information Sciences,
Pakistan Institute of Engineering and Applied Sciences,
Islamabad, Pakistan
alirazaghuman33@yahoo.com

Hafiz Saud Arshad
Dept. of Computer and Information Sciences,
Pakistan Institute of Engineering and Applied Sciences,
Islamabad, Pakistan
saudarshad25462@gmail.com

*Abstract*—**There are numerous peptides discovered through past decades, which exhibit antimicrobial and anti-cancerous tendencies. Due to these reasons, peptides are supposed to be sound therapeutic candidates. Some peptides can pose low metabolic stability, high toxicity and high hemolity of peptides. This highlights the importance for evaluating hemolytic tendencies and toxicity of peptides, before using them for therapeutics. Traditional methods for evaluation of toxicity of peptides can be time-consuming and costly. In this study, we have extracted peptides data (Hemo-DB) from Database of Antimicrobial Activity and Structure of Peptides (DBAASP) based on certain hemolity criteria and we present a machine learning based method for prediction of hemolytic tendencies of peptides (i.e. Hemolytic or Non-Hemolytic). Our model offers significant improvement on hemolity prediction benchmarks. we also propose a reliable clustering-based train-tests splitting method which ensures that no peptide in train set is more than 40% similar to any peptide in test set. Using this train-test split, we can get reliable estimated of expected model performance on unseen data distribution or newly discovered peptides. Our model tests 0.9986 AUC-ROC (Area Under Receiver Operating Curve) and 97.79% Accuracy on test set of Hemo-DB using traditional random train-test splitting method. Moreover, our model tests AUC-ROC of 0.997 and Accuracy of 97.58% while using clustering-based train-test data split. Furthermore, we check our model on an unseen data distribution (at Hemo-PI 3) and we recorded 0.8726 AUC-ROC and 79.5% accuracy. Using the proposed method, potential therapeutic peptides can be screened, which may further in therapeutics and get reliable predictions for unseen amino acids distribution of peptides and newly discovered peptides.**

## I. Introduction

Drug discovery and synthesis have been two of the most active research areas in bio-informatics since past few years. This is a very promising field with direct impact to vaccine development and therapeutic industries. Therapeutic industry has been using different protein and peptide antibodies in response to antigens. Antigens are the alien substances that can stimulate an immune response (specifically activating lymphocytes – which are infection fighting white blood cells). Usually as a result of immune response, antibodies are generated by the immune system that fight with the infection causing substance. Advancements led to the realization that antibodies and a major group of Antigens are proteins, and different types of antibodies are supposed to be directed to different proteins and peptides. This realized a need for peptide antibodies.

Therapeutics usually use large protein sequences and peptide antibodies. Both proteins and peptides are made up off Amino acids held together by peptide bonds. The major difference being that peptides are composed of smaller amino acid sequences as compared to proteins. Although in recent years, due to the inherent limitations of protein antibodies such as large size, low specificity, insufficient affinity and inefficiency towards toxic or hazardous proteins, researchers have realized the need to further the search for other options. Therapeutics are now looking for comparatively smaller molecules with high specificity, high affinity, better tissue penetration, low production cost and ability to attach to toxic proteins; although not a big success yet. Based on this, peptides have gained much importance. Peptides were crucial reagents for elucidating the molecular biology of antibody specificity and biosynthesis. Peptides have shown antimicrobial, antimalarial, anti-parasitic, anti-cancerous and cell penetrating tendencies which are very favorable for therapeutics. Peptides and antibodies have entered a fruitful companionship in immunology since they were discovered.

Despite all above merits, the number of therapeutic peptides drugs that hit the market are not significant as projected. Few major reasons are low metabolic stability, high toxicity and high hemolysis of peptides. Previously, there's been some work done on the detection of toxicity in peptides i.e. classified usually into hemo-toxicity (lysis of red blood cells), cypo-toxicity and immunotoxicity. High hemolytic peptides can cause dead red blood cells which is not favorable for therapeutics. For a drug's success, it is necessary that peptides possess no or low hemolytic tendencies. This study is based on prediction of hemo-toxicity of peptides.

Hemo-toxicity of peptides can be predicted in different ways. One way of predicting hemo-toxicity of peptides is to identify and analyze different motifs and amino acid chains in already known hemotoxic peptides, and then search the test sample for these hemotoxic motifs. If a peptide contains motifs usually found in hemolytic peptides then peptides is classified as hemolytic. This analysis can be very unreliable, time consuming and cumbersome for proteins of much longer chains of amino acids. Another option could be to have the set of all known hemotoxic proteins and use sequence alignment tools to check the similarity between the test protein sequence and known hemotoxic protein sequences. If similarity in specific motifs passes a threshold then the peptide is considered hemotoxic. This technique also doesn't promise reliable results and may require too much of parameter settings, which can be cumbersome too. Thus, one must look for an automated method that could get faster results with lesser efforts. This brings us to using machine learning methods for this task. Different machine learning models are trained with known hemotoxic and non-hemotoxic sequences, and later these models can be used for prediction of hemo-toxicity for peptides.

## II. LITERATURE REVIEW

Recently, researcher have turned to investigate why therapeutic peptides do not make it to clinical trials. It has been observed some chemotherapeutic peptide drugs induce hemolysis which can hinder antimicrobial activities[1]. So, Studies have concluded that High Hemolytic peptides are not suitable for clinical trials[2]. In order to categorize a peptide to be highly hemolytic, we need some criterion. One generally used criterion[3] is known as HC-50 (defined as Hemolytic Concentration of peptides to obtain 50% of lysis of erythrocytes). Some studies[4], [5] also apply two additional conditions other than HC-50 for a peptide to be classified as highly hemolytic. There have been efforts made to predict the hemolytic tendency of natural peptides[4], [6], [7], chemically and structurally modified peptides[5]. Peptide sequences cannot be processed directly through a predictor function (which would predict the hemolytic potency of the peptide). Hence, peptides are represented in a form that is cohesive with predictor function's input. This representation is called features of the peptide. Different studies extracted different types of features like Amino Acid Composition[4]–[6], Dipeptide Composition[4]–[6] and Physicochemical properties[4]–[6]. Some researchers have also used structure-based descriptors[8]. Another prior work[5] has employed feature selection[9] before feeding the feature matrix to machine learning method. Many studies[4]–[6] have employed the classical machine learning methods like Support Vector Machines[10], Decision Trees, Random Forest and K-Nearest neighbor.

## III. METHODOLOGY

We are to develop machine learning models to predict the hemo-toxicity of peptides. Machine learning models can be of two different types namely supervised and unsupervised. Unsupervised machine learning techniques are the ones which do not require labels (e.g. hemolytic or non-hemolytic) for a specific data sample, in order to develop a prediction model. The other way to go is supervised machine learning methods which require a label for each data sample, in order to learn to classify a sample between different classes (e.g. hemolytic or non-hemolytic). We will proceed with supervised methods as they tend to be more reliable while testing.

In order to develop a ML model, usually a typical ML pipeline is to be followed. First of all, a database is required (with known example-label pairs). This database consists of sequences of amino acids and all sequences can be of different lengths. Machine Learning models are trained on fixed length numeric data. For the purpose, we convert each amino acid sequence to a fixed length vector of counts of different amino acids. These data samples (vectors) are pre-processed to remove biases in the data and identification of outliers. After that, data is divided into two parts, namely train set and test set. Following that, different ML models are to be trained on the train-set. Finally, model's performance is tested on test set. If the performance is acceptable, model is selected and deployed. If performance is not good enough, model is retrained and re-evaluated.

A typical machine learning model development pipeline is shown in the below diagram.

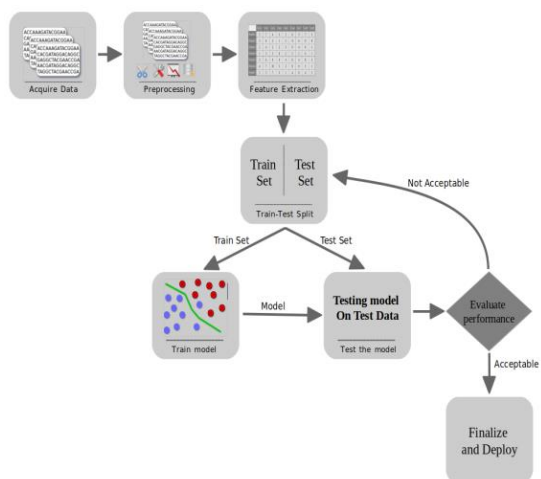

Fig-1: A typical Machine Learning pipeline

### A. Acquiring Data:

Peptides are extracted from different experimentally validated databases. These databases are vast collections of peptides collected from different subjects compiled by different labs around the world. These peptide samples comprise of natural Amino Acids. These databases represent the hemolytic tendencies e.g. 10% hemolysis or 60% hemolysis of peptides along with their corresponding concentration. We need a criterion to determine what minimum concentration of hemolysis can proclaim a peptide to be hemolytic. Although, this criterion must depend on what therapeutics are to be synthesized but for benchmarking and standardization purposes of different methods, there are few standard criteria that are used. The criterion used by Chaudhary et. al.[4] is usually used as a standard for benchmarking purposes which is also used by later works like Kumar et al.[5]. Chaudhary et al.[4] extracted database is called Hemo-PI 3 database. We use Hemo-PI 3 for benchmarking of our methods and extract another dataset based on same criteria as Hemo-PI 3. The criteria are shown in Table-I.

TABLE I. TABLE TYPE STYLES

| Hemolytic *Peptides* | | Non-Hemolytic peptides | |
|---|---|---|---|
| Hemolysis | Concentration | Hemolysis | Concentration |
| ≥ 5% | ≤ 10 | ≤ 2 % | ≥ 10 |
| ≥ 10% | ≤ 20 | ≤ 5 % | ≥ 20 |
| ≥ 15% | ≤ 50 | ≤ 10 % | ≥ 50 |
| ≥ 20% | ≤ 100 | ≤ 15 % | ≥ 100 |
| ≥ 30% | ≤ 200 | ≤ 20 % | ≥ 200 |
| ≥ 50% | ≤ 300 | ≤ 30 % | ≥ 300 |
| | | ≤ 50 % | ≥ 500 |

Peptides were quantified as hemolytic or non-hemolytic based on above given criterion. The ones that don't meet this criterion are considered as non-hemolytic in nature. We chose and extracted peptides on above given criterion from database called DBAASP[11]. We name this collected dataset as Hemo-DB. Hemo-DB database comprises of 3268 experimentally validated peptides based on their hemolysis. Hemo-DB dataset is specifically designed for purpose of discriminating between hemolytic and non-hemolytic peptides. 1720 of these peptides are hemolytic (strongly hemolytic) and rest of the peptides are non-hemolytic (Poorly Hemolytic) as per criteria mentioned in Table-I. While Hemo-PI 3 which is used for benchmarking consists of 1623 experimentally validated peptides out of which 686 peptides are extracted from DBAASP[11] and rest 937 are extracted from database Hemolytik[12].

*B. Feature Extraction:*

Extracted data (Hemo-DB) is in the form of sequences of amino-acids i.e. peptides. Despite that, these sequences are of different lengths. These peptides cannot be directly processed through a machine learning (ML) pipeline as ML pipeline works on numerically represented fixed length data. So, we transform the data into a mathematical representation. All amino acid frequencies are calculated and stored as a representation for the corresponding peptide sequence. Now each peptide sequence is stored as a vector of counts of amino acids in the sequence. These frequencies are fixed length vectors of numbers which can be processed through ML pipeline.

Following is a t-SNE (t - Distributed Stochastic Neighbor Embedding) transform of the dataset. Fig-2 shows data samples 20-dimensional dataset transformation to 2D.

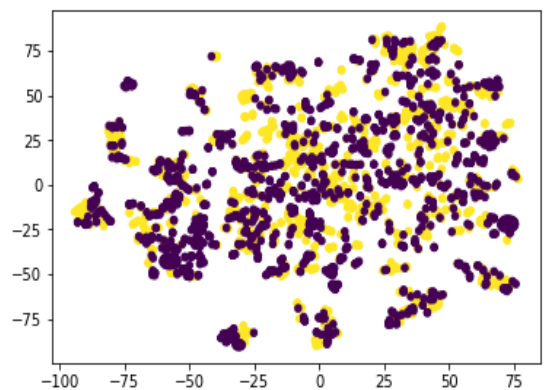

Fig-2: t-SNE transform visualization of data samples in database

As we can see that data sample are densely populated in feature space and one can deduce that these samples cannot be discriminated using a linear classification model. Furthermore, we can visualize the importance of features using scree-plot. Fig-4 shows a Scree-plot which plots cumulated variance captured by features (amino acid counts).

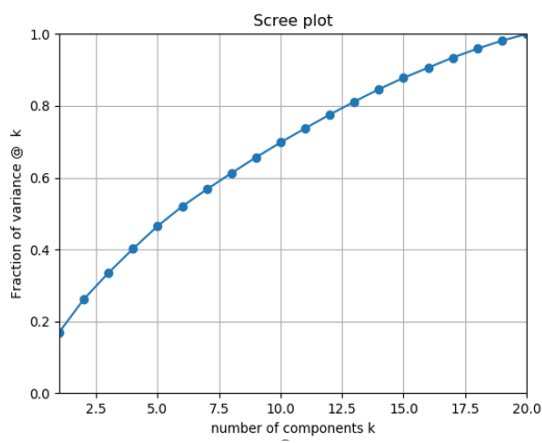

Fig-3: Scree-plot showing captured cumulated variance by number of features (components of feature vector)

*C. Pre-processing:*

After feature extraction, each peptide sequence (sample) is an N(number of unique amino-acids) dimensional vector of frequencies. The extracted feature vector (amino-acid frequencies) of peptides can be over-whelmed by frequency of a few amino-acids and the rest may lose their significance in the representation of peptides. So, we normalize all these vectors to unit norm, so that learning is not hijacked by some specific amino-acid frequencies. All frequencies are normalized in a way that frequencies differ in their values so that their significance is still preserved, yet higher frequency amino-acids don't spike so high to hijack the learning process. In this way, each feature has a unit weight and all data samples have equal contribution in learning process as well.

*D. Clusters:*

Usually peptides are randomly distributed in train and test sets for machine learning approaches regardless of the features of peptides. These methods assume that test peptides and train peptides have the same distribution of Amino-Acids. Tends to not perform reliably for a newly discovered or synthesized peptides, having significantly different distribution of Amino-acids than the distribution of training peptides.

To handle this discrepancy, we propose to cluster peptides based on their amino acids sequence similarities. For example, each peptide will share a minimum of 40% similarity with all other peptides in the same cluster while share less than 40% similarity with peptides from all other clusters. This way we have sets of similar peptides represented by each cluster. Clustering was carried in a manner that we performed clustering separately on both mutually exclusive sets i.e. hemolytic and non-hemolytic peptides from Hemo-DB. Clustering gives us a total of 300 clusters. 135 of the total clusters are hemolytic and 165 are non-hemolytic. It is to note that originally the dataset had more hemolytic peptides that non-hemolytic and after clustering total hemolytic peptides clusters are less than non-hemolytic clusters. This tells that hemolytic peptides show more similarity among each other as compared to non-hemolytic peptides.

*E. Train-Test Sets:*

Some of the clusters formed after clustering step were dedicated for training and rest for testing. This way the training set and test set are mutually exclusive and share least similarity with each other. A model which performs well on testing set (clusters), is also expected to perform reliably on an unseen or newly discovered peptide.

For dataset partitioning, we did 10 folds cross-validation. This partitions the dataset into 10 equal sized partitions of clusters. We did stratified sampling from both hemolytic and non-hemolytic clusters because out of the original database, peptide clusters were not equally distributed. This ensures unbiased training. At 1st fold, let's say first 9 partitions go into train set and the 10th partition goes into test set. A model is trained on train set, tested on test set and performance is recorded. Next time, 9th partition be picked for test set and rest go to train set. A model is trained and performance on test set is recorded. At last, mean of performances of all folds is reported as the expected performance of the model.

*F. Model Development:*

There are different types of machine learning models that can be trained on the extracted features of the peptides. We've experimented with different categories of machine learning algorithms namely Random Forest, Support Vector Machines and XG-Boost. Random Forest consists of a set of decision trees, which partition the sample space into smaller partitions until all the training samples are classified into right classes. Main parameters of Random forest algorithms are maximum depth of any tree and number of decision trees. The second algorithm is support vector machines (SVM). SVM's basic working principal is to maximize the margin (distance) of discriminating hyperplane from internal boundary points of the two classes examples.

The capacity of SVM is extended with non-linear kernels or kernel tricks to solve much more complex problems like highly non-linear feature space and heterogeneous feature types. Another algorithm used in this study is XG-Boost. XG-Boost is basically a boosting overlay on multiple weak learners (decision trees). This method has been mostly adopted for different data science studies because it is not a black box model and provides insights into what criterion is used by the model to classify an example.

## IV. EVALUATION AND COMPARISON

For first, we applied different machine learning models on the peptide feature vectors with traditional 10 fold cross validation and stratified sampling, and observed how it performs on pristine features. We recorded 0.73 mean specificity, 0.667 mean sensitivity, 70% mean accuracy and 0.7706 mean Area Under ROC using Support Vector Machines (SVM). These were the results on best tuned parameters of SVM. Despite that we have also applied Random Forest and Boosting Techniques like XG-Boost. Results are shown in the following Table-II.

TABLE II. MODEL EVALUATION ON HEMO-DB WITHOUT FEATURE NORMALIZATION AND CLUSTERING

| Models | Specificity | Sensitivity | Accuracy | Area Under Curve |
|---|---|---|---|---|
| SVM | **0.7302** | 0.6679 | **0.7007** | **0.7706** |
| Random Forest | 0.7011 | **0.6770** | 0.6897 | 0.7509 |
| XG-Boost | 0.7087 | 0.6498 | 0.6808 | 0.7352 |

If we normalize the feature vectors to unit norm, results improve quite significantly. Improved performance of three aforementioned ML models are shown in Table-III.

TABLE III. MODEL EVALUATION ON HEMO-DB AFTER CLUSTERING AND FEATURE NORMALIZATION

| Models | Specificity | Sensitivity | Accuracy | Area Under Curve |
|---|---|---|---|---|
| SVM | 0.9034 | 0.1304 | 0.5373 | 0.5404 |
| Random Forest | 0.9441 | 0.9689 | 0.9559 | 0.9911 |
| XG-Boost | **0.9744** | **0.9819** | **0.9779** | **0.9986** |

Now, we split the data based on proposed clustering-based method and apply k-fold cross validation on peptides of training algorithm for reliable evaluation. Following Table-IV shows results of these models on testing clusters of peptides. We obtained mean 0.997 AUR-ROC with 8.18e-05 variance.

TABLE IV. MODEL EVALUATION ON HEMO PI 3 TEST SET

| Models | Specificity | Sensitivity | Accuracy | Area Under Curve |
|---|---|---|---|---|
| SVM | 0.2 | 0.8010 | 0.4847 | 0.4550 |
| Random Forest | 0.9313 | 0.8817 | 0.9078 | 0.9788 |
| XG-Boost | **0.9767** | **0.9748** | **0.9758** | **0.9970** |

It can be observed that after normalization tree-based methods tend to perform better and consistent. We also notice that after clustering, accuracy dropped significantly on models like SVM and Random Forest. XG-Boost performance is somewhat consistent because boosting methods use multiple weak learners to predict hemolity of peptide sequence. Despite that, following is the ROC curve for XG-Boost with normalized feature vectors for Hemo-DB.

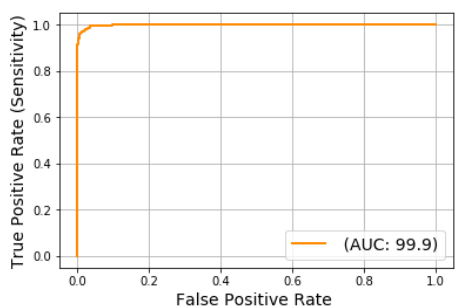

Fig-4: ROC curve for XG-Boost model with normalized feature vector

As a benchmark and evaluation of model on unknown distributions of data, we test our XG-Boost model on Hemo-PI 3 database as used by Chaudhary et al.[4]. Following plot shows the ROC using XG-Boost model on Hemo-PI 3 database.

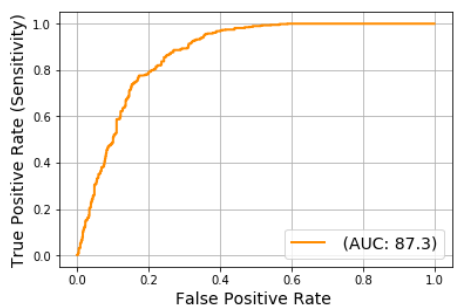

Fig-4: ROC curve for XG-Boost model with normalized feature vector for Hemo-PI 3

Testing the model on Hemo-PI 3, we get 0. 8726 mean Area under ROC, 79.5% mean accuracy, 0.8474 mean sensitivity and 0.7322 mean specificity using XG-Boost model trained on Hemo-DB. We can observe that the model does not perform on Hemo-PI3 as good as on Hemo-DB database; yet, well enough, to be considered a reliable and consistent model. These results already surpass the prior results[4] without being trained on data from the distribution of Hemo-PI 3 database. This is a solid indicator of reliability and consistency of the model.